# Predicting Risk-of-Readmission for Congestive Heart Failure Patients: A Multi-Layer Approach

Kiyana Zolfaghar, MS[1], Nele Verbiest, MS[2], Jayshree Agarwal, MS[1], Naren Meadem, BS[1], Si-Chi Chin, PhD[1], Senjuti Basu Roy, PhD[1], Ankur Teredesai, PhD[1], David Hazel, MS[1], Paul Amoroso, MD[3], Lester Reed, MD[3]

[1]Institute of Technology, CWDS, University of Washington Tacoma, WA; [2]Department of Applied Mathematics, Computer Science and Statistics, Ghent University, Belgium; [3]Multicare Health System, Tacoma, WA

**Abstract**

*Mitigating risk-of-readmission of Congestive Heart Failure (CHF) patients within 30 days of discharge is important because such readmissions are not only expensive but also critical indicator of provider care and quality of treatment. Accurately predicting the risk-of-readmission may allow hospitals to identify high-risk patients and eventually improve quality of care by identifying factors that contribute to such readmissions in many scenarios. In this paper, we investigate the problem of predicting risk-of-readmission as a supervised learning problem, using a multi-layer classification approach. Earlier contributions inadequately attempted to assess a risk value for 30 day readmission by building a direct predictive model as opposed to our approach. We first split the problem into various stages, (a) at risk in general (b) risk within 60 days (c) risk within 30 days, and then build suitable classifiers for each stage, thereby increasing the ability to accurately predict the risk using multiple layers of decision. The advantage of our approach is that we can use different classification models for the subtasks that are more suited for the respective problems. Moreover, each of the subtasks can be solved using different features and training data leading to a highly confident diagnosis or risk compared to a one-shot single layer approach. An experimental evaluation on actual hospital patient record data from Multicare Health Systems shows that our model is significantly better at predicting risk-of-readmission of CHF patients within 30 days after discharge compared to prior attempts.*

**Introduction**

With the overwhelming increase in available health care data, analyzing and mining this data has gained more interest over the last decade. Improving awareness, personalizing medical treatments and ameliorating health care standards are only a few examples of opportunities that result from mining health care data[1].

In this work, we focus on building a predictive model to enhance quality of care[2] for patients with cardiac heart failure. The main goal is to predict the level of risk of patients being discharged after a Congestive Heart Failure (CHF) in order to assess if they are likely to be at high risk of readmission within the next 30 days. We approach this as a *classification problem to classify patients into high or low risk given historical discharge history data along with variety of other parameters*. We leverage historic patient data that contains admission-readmission histories of CHF patients. . Moreover, hospital readmission is expensive and generally preventable[3]. If CHF readmission could be predicted accurately, hospitals would invest more purposefully in improving hospital care by reducing risk of infection, reconciling medications, educating patients on what exact symptoms to monitor, and assess readiness of patients for discharge[4]. At first, the 30 day window seems to be arbitrary, but it is indeed a clinically meaningful time window for hospitals, and the Center for Medicare and Medicaid Services (CMS) has started using the 30 day all cause heart failure readmission rate as a publicly reported efficiency metric. Moreover, all cause 30 day readmission rate for patients with CHF has increased by 11 percent between 1992 and 2001[15].

Predicting if patients discharged with CHF will be readmitted within 30 days is traditionally approached as a single classification task. We observe two main drawbacks of this approach: (a) firstly, classification of risk of readmission is highly imbalanced, as can be seen from Figure 1, and is hence inherently difficult to solve[5], and (b) secondly, (COMPLETE THIS HERE) Traditional classification methods will generally tend to assign most of the patients to the majority class (no readmission), as the training data consists mostly of majority instances. Another issue is in including all patients discharged with CHF to build the classification model might not be meaningful, as patients that were discharged after a long length of stay can have characteristics that are totally different from patients that were discharged after a short length of stay , and are hence irrelevant for the 30 days classification task.

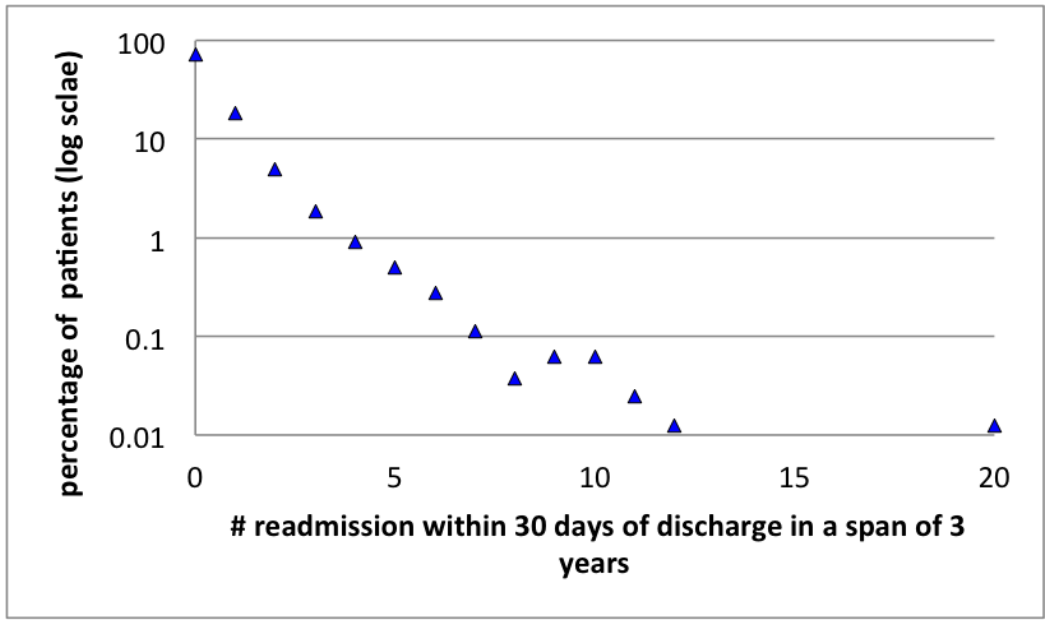

**Figure 1: The number of times a patient was readmitted within 30 days after discharge from CHF in a span of 3 years.**

In this paper we address these drawbacks by introducing a multi-layer classification strategy. The main idea is: we first build a rough model that predicts if patients will be readmitted within a given time window longer than 30 days, and then use a more refined model to predict if patients will be readmitted within 30 days. Specifically, in order to predict if any patient discharged after CHF will be readmitted within 30 days, we first use a coarse grain model to predict if the patient is likely to be readmitted at all (in any reasonable timeframe). If not, we can mostly conclude that the patient will not be highly likely to be readmitted within 30 days (a very short timeframe). Else, we predict if the patient will be readmitted within a large time window. If not, than we can conclude that the patient will not be readmitted within 30 days. If the outcome is that the patient will be readmitted within the large time window, we can use the more refined model to predict if the patient will be readmitted within 30 days.

This multi-layer classifier allows for flexibility in many ways. The main advantage is that we can use different models for respective granularity of problems. If we use different classifiers for different layers, we can use different features for each layer; and the classification tasks can be more refined as it only considers patients in the training data that were readmitted within the large time window. The second advantage is that we can split up the imbalanced classification problem in two more or relatively more balanced classification problems.

The main contributions of this paper are:

- We introduce a multi-layer classifier to predict if patients are likely to be readmitted within 30 days after being discharged from CHF
- We perform an experimental study using a real-world data set provided by the Multicare Health Systems

The remainder of this paper is structured as follows. In the next Section, we describe our multi-layer approach in detail, and describe the classifiers and feature selection methods that are used in the layers. Next, we evaluate the performance of our approach in the experimental Section, and compare it with state-of-the-art methods. Afterwards, we study related work, and we conclude and suggest further research directions in the concluding Section.

**Multi-layer Classification for Readmission of Congestive Heart Failure Patients**

In this section we propose a multi-layer classifier method for predicting readmission of congestive heart failure patients. Instead of tackling the classification problem at once, we divide it in three sub-problems, as depicted in Figure 2. For a new patient discharged after CHF treatment, we first predict if she will ever be readmitted to the hospital. If the prediction is that the patient will likely never be readmitted, we are done with the prediction task. If the outcome is that the patient may be readmitted (i.e. predicted yes), we use another model (layer) to predict if the patient will be readmitted within 60 days. Again, if the outcome is no, this means that the patient will not be readmitted within 60 days, and hence we output that the patient will not be readmitted within 30 days neither. If the outcome is again a yes, we use yet another model (hence multi-layer) to predict if the patient will be readmitted within 30 days. The outcome of this final classification is then returned as the final classification.

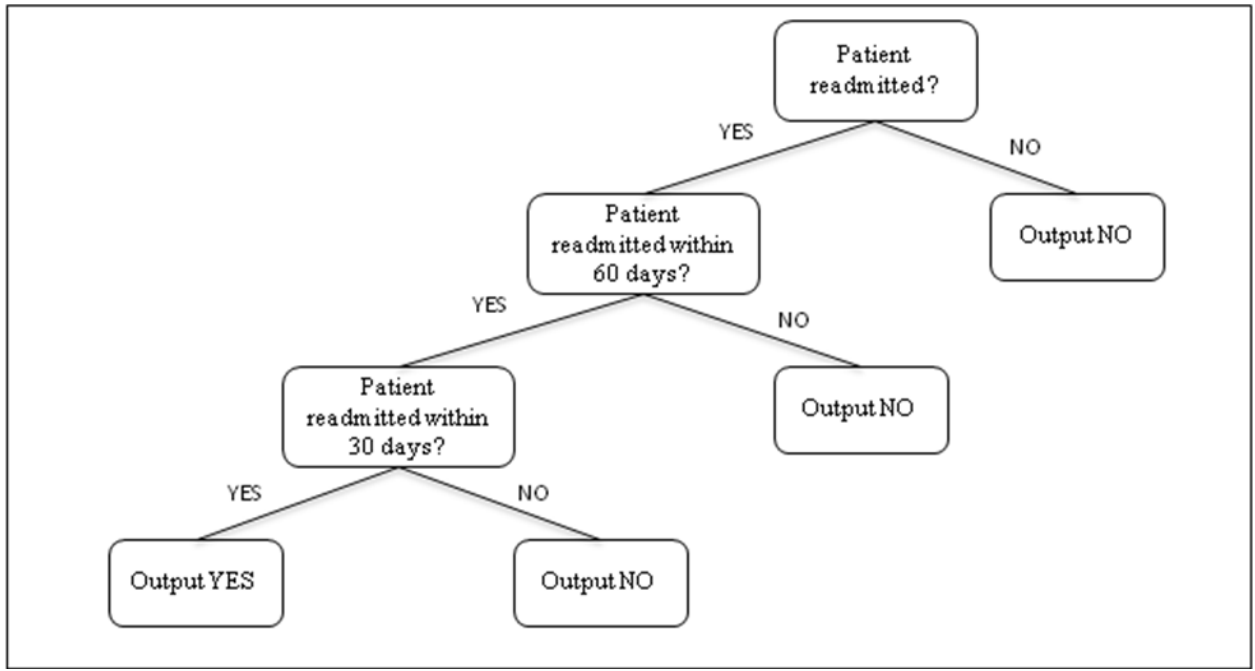

**Figure 2 Subdivision of the classification problem into multiple layers.**

Training data that is used in each layer is different. The upper layer uses all the training data. At the second layer, only the patients in the training data that are readmitted are used. In the last layer of the problem, only the patients that are admitted within 60 days are used. As a result, the training data that is used in the second and final layer is more refined than the original data. The purpose of this is to provide each sub-problem only with the relevant data. For example, if we want to predict if a patient will be readmitted within 30 days, the information about patients that will never be readmitted is not relevant and might disturb the classification.

Another important advantage of this approach is that the highly imbalanced problem is divided into three more or less balanced problems. The data distribution is depicted in Figure 3. In general, a classification problem is called imbalanced if its Imbalance Ratio (IR, number of majority instances divided by the number of minority instances) is more than 2. In the original problem, the positive class (patients readmitted within 30 days) covered 1477 patients, while the majority class covered 8293 patients. The imbalance ratio of this problem is 5.6, making it severely imbalanced. Number of patients that was never readmitted is 5503 and the total number of patients considered is 9770, resulting in an IR of 1.7 leading to a more balanced problem that is generally easier to solve. The threshold 60 at the second layer of the multi-classifier was chosen to balance the second layer problem, such that the IR of the second layer is 1. The number of patients that were readmitted within 30 days is 1477, so the IR of the final layer is 1.4. We conclude that using this multi-layer approach, the heavily imbalanced original problem is divided into subtasks (layers) that are more or less balanced.

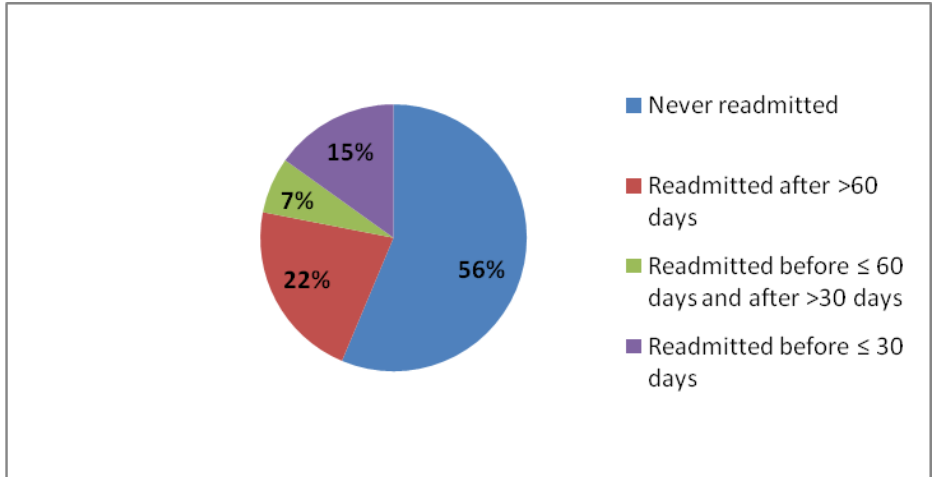

**Figure 3: Distribution of the patients based on the number of days until readmission after CHS. By dividing the problem in three parts, each of the subtasks is balanced.**

Furthermore, we can consider different features in each sub-problem. For instance, features which are good to predict if a patient will ever be readmitted or not, might not be relevant features to predict if the patient will be readmitted within 30 days. Therefore, we apply feature selection in every layer of the multi-layered classifier. As a result, each layer will work with features that are suited for the corresponding classification task.
The feature selection technique that we use in this paper is the Chi-square test[6], as this technique has proven to be successful in earlier works. This test calculates for each feature a score that expresses its relevance with respect to the decision class, and then decides based on this score which features to retain.

Finally, we can also use different classifiers for the different sub-problems. There are two advantages related to this property. The first one is that it can occur that one classifier is well suited for one classification problem but not for the other. For instance, one classifier can work well for the second layer problem, but not for the third layer problem. Secondly, some classifiers require a longer running time than others, and it might not always be feasible to apply them to each layer of the problem. However, it is possible to apply these more involved classifiers to the final layer of the classification problem. We hope that using a more refined classifier for the final layer of our approach will improve classification results.

We propose two different multi-layer classifiers, as described in Table 2. The first classifier, to which we will refer to as MLC1, is a multi-layer classifier that uses the Naïve Bayes (NB[7]) classifier in each layer of the problem. The second classifier, called MLC2, uses NB in the first two coarse layers of the problem, and then uses a Support Vector Machine (SVM[8]) classifier for the final classification problem.
We work with NB because it is a fast and simple model that has shown to be effective in many real-world problems. The SVM classifier is more time-consuming, but it is generally more accurate. Therefore, we use it in the last layer of one of the multi-layer classifiers.

**Table 1: The classifiers (NB or SVM) that are used in each layer of the two multi-layer classifiers.**

|  | MLC1 | MLC2 |
|---|---|---|
| Predicting if patient will be ever readmitted | NB | NB |
| Predicting if patient will be readmitted within 60 days | NB | NB |
| Predicting if patient will be readmitted within 30 days | NB | SVM |

**Experimental Evaluation: Set-up**

The dataset used to derive our readmission prediction model is provided by Multicare Health System (MHS). We are given a set of tables where each table contains data related to the patients. Hospital encounters with discharge diagnosis of CHF (primary or secondary) are considered as the potential index admission due to CHF. We only consider patients with a discharge diagnosis of the International Classification of Diseases, 9th Revision, Clinical Modification Codes (ICD-9 CM) related to CHF, listed in Table 2.

**Table 2: The ICD-9 CM codes for CHF**

| ICD-9 CM codes | Description |
| --- | --- |
| 402.01 | Malignant hypertensive heart disease with heart failure |
| 402.11 | Benign hypertensive heart disease with heart failure |
| 402.91 | Unspecified hypertensive heart disease with heart failure |
| 404.01 | Malignant hypertensive heart and kidney disease with heart failure and with chronic kidney disease stage I through stage IV, or unspecified |
| 404.03 | Malignant hypertensive heart and kidney disease with heart failure and chronic kidney disease stage V or end stage renal disease |
| 404.11 | Benign hypertensive heart and kidney disease with heart failure and with chronic kidney disease stage I through stage IV, or unspecified |
| 404.13 | Benign hypertensive heart and kidney disease with heart failure and chronic kidney disease stage V or end stage renal disease |
| 404.91 | Unspecified hypertensive heart and kidney disease with heart failure and with chronic kidney disease stage I through stage IV, or unspecified |
| 404.93 | Unspecified hypertensive heart and kidney disease with heart failure and chronic kidney disease stage V or end stage renal disease |
| | |
| 428.XX | Heart Failure codes |

All the patients can be identified by a unique patient id and each hospital encounter is uniquely identified by an admission id. Multiple admissions (i.e., readmissions) of the same patient can be identified by using the patient id. Our entity of observation is each CHF hospital encounter and we consider only the admissions when a patient is discharged to home to exclude inter hospital transfers. Admissions encountering in-hospital deaths are not included in our analysis because we are more interested in predicting readmissions. We calculate the days elapsed between the last discharge due to CHF and next admission in order to identify if the readmission has occurred within 30 days. The dataset consists of CHF hospitalization for patients discharged since 2009. It provides information of 6348 patients diagnosed with CHF and number of hospital encounters generated by these patients during 2009-2012 is 11383. As mentioned earlier, various supporting tables are provided to get a complete understanding the patients related to heart failure and to identify the attributes to be used as predictor variables in modeling. The detailed description of some of the attributes is given in Table 3.

The key socio-demographic factors related to patients are, gender, race, marital status. Some of the other important factors pertinent to CHF are ejection fraction which represents the volumetric fraction of blood pumped out of the ventricle with each heartbeat, blood pressure, primary and secondary diagnosis, other comorbidity variables, APR-DRG code (All Patient Refined Diagnosis Related Groups Definition; a classification system that classifies patients according to reason of admission) for severity of illness and APR-DRG code for risk of mortality. Information about the discharge disposition of patients like the discharge status, discharge destination, length of stay and follow-up plans are also found to be correlated to CHF readmissions. In addition, 34 cardiovascular and comorbidity attributes[14] mentioned in Table 3 are also used. Based on our initial understanding we observed that ejection fraction has about 59% of missing values followed by APR-DRG code for severity of illness (13.3%) and blood pressure (12.6%). We imputed the missing value of ejection fraction and after removing the instances with other null values; our final dataset consists of 9770 instances on which the model is built.

**Table 3:** Description of different attributes

| Variable | Type | Mean/No. of Domain Values |
|---|---|---|
| Age | Numeric | 69 |
| Gender | Categorical | 2(M, F) |
| Marital status | Categorical | 9 such as married, divorced |
| Ethnic group | Categorical | 9 such as Caucasian, Asian , African-American |
| Discharge follow-up plan | Categorical | 7 such as 2 days, 5 days |
| Discharge destination | Categorical | 70 |
| Discharge status | Categorical | 15 such as discharged to home, discharged to rehab facility |
| Admit source | Categorical | 6 such as transfer from hospital, emergency room |
| Admit type | Categorical | 4 such as elective, emergency |
| Blood Pressure | Categorical | 9 |
| Ejection fraction value | Numeric | 48.63 |
| Secondary diagnosis count | Numeric | 16.56 |
| Discharge APR-DRG Severity of illness | Categorical | 4 such as 1(least severe), 2, 3, 4(most severe) |
| Discharge APR-DRG Risk of mortality | Categorical | 4 such as 1(least severe), 2, 3, 4(most severe). |
| Length of stay | Numeric | 5 |
| IsHFPrimary | Categorical | 2(Y,N) |
| Congestive heart failure | Categorical | 2 (0,1) |
| Acute coronary syndrome | Categorical | 2 (0,1) |
| Arrhythmias | Categorical | 2 (0,1) |
| Cardio-respiratory failure and shock | Categorical | 2 (0,1) |
| Valvular and rheumatic heart disease | Categorical | 2 (0,1) |
| Vascular or circulatory disease | Categorical | 2 (0,1) |
| Chronic atherosclerosis | Categorical | 2 (0,1) |
| Other and unspecified heart disease | Categorical | 2 (0,1) |
| Hemiplegia, paraplegia, paralysis, functional disability | Categorical | 2 (0,1) |
| Stroke | Categorical | 2 (0,1) |
| Renal failure | Categorical | 2 (0,1) |
| COPD | Categorical | 2 (0,1) |
| Diabetes and DM complications | Categorical | 2 (0,1) |
| Disorders of fluid/electrolyte/acid base | Categorical | 2 (0,1) |
| Other urinary tract disorders | Categorical | 2 (0,1) |
| Decubitus ulcer or chronic skin ulcer | Categorical | 2 (0,1) |
| Other gastrointestinal disorders | Categorical | 2 (0,1) |
| Peptic ulcer, hemorrhage, other specified gastrointestinal disorders | Categorical | 2 (0,1) |
| Severe hematological disorders | Categorical | 2 (0,1) |
| Nephritis | Categorical | 2 (0,1) |
| Dementia and senility | Categorical | 2 (0,1) |
| Metastatic cancer and acute leukemia | Categorical | 2 (0,1) |
| Cancer | Categorical | 2 (0,1) |
| Liver and biliary disease | Categorical | 2 (0,1) |
| End-stage renal disease or dialysis | Categorical | 2 (0,1) |
| Asthma | Categorical | 2 (0,1) |
| Iron deficiency and other/unspecified anemias and blood disease | Categorical | 2 (0,1) |
| Pneumonia | Categorical | 2 (0,1) |
| Drug/alcohol abuse/dependence/psychosis | Categorical | 2 (0,1) |
| Major pysch disorders | Categorical | 2 (0,1) |
| Depression | Categorical | 2 (0,1) |
| Other psychiatric disorders | Categorical | 2 (0,1) |
| Fibrosis of lung and other chronic lung disorders | Categorical | 2 (0,1) |
| Protein-calorie malnutrition | Categorical | 2 (0,1) |

We compare our model with two relevant baseline methods. Both baseline methods first apply the same feature selection method to the data as in our model, namely Chi-Square. After that, we use both NB and SVM to classify the data. Both baseline methods use all the data to predict if a patient discharged from CHS will be readmitted within 30 days.

Before running the algorithms on the data, we first impute missing values in the Ejection Fraction feature. We do this both for the baseline methods as for our proposed method. The instances that have missing values in other features are removed from the dataset. As we do this for both the baseline methods and our proposed multi-layer classifier, we obtain a fair comparison. The reason why we only impute the missing values in the Ejection Fraction feature is that this feature has a high percentage of missing values (about 60 percent) and that this approach has proven to work well in preliminary experiments[9].

We perform a 10 fold cross validation procedure, that is, the data is divided into 10 equal folds, and each fold is considered as test data, that is classified using a model that is built on the remaining 9 folds, called the training data. As each fold is considered once as test data, we obtain one single classification outcome for each instance in the set.

The outline of the experiments is depicted in Figure 4.

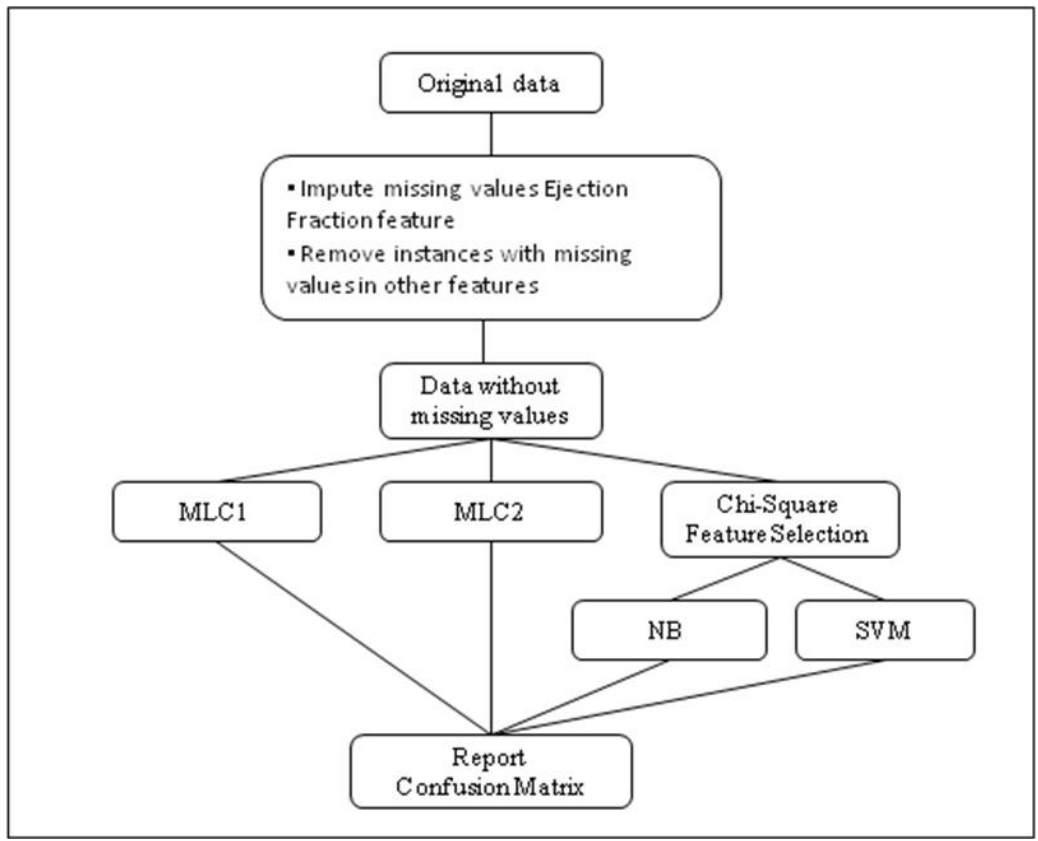

**Figure 4: Structure of the experimental set-up**

**Experimental Evaluation: Results**

In this section, we present and discuss the results obtained with our multi-layer classification approach, and compare it to the baseline approaches. In Table 2, we show the confusion matrix values for all methods. The positives refer to the patients that are readmitted within 30 days to the hospital after discharge from CHS, while the negatives refer to all other patients. For instance, True Positives refers to the patients that were readmitted within 30 days to the hospital, and that were also predicted by the respective classifier to be readmitted within 30 days. On the other hand, False Negatives refers to patients that were readmitted within 30 days to the hospital, but for which the classifier predicted that the patient would not be readmitted within 30 days. These numbers give a good insight in the performance of the classifier, especially because the considered problem is highly imbalanced. Only reporting accuracy would give a false image of the results.

|  | True Positives (TN) | False Positives (FP) | True Negatives (TN) | False Negatives (FN) |
| --- | --- | --- | --- | --- |
| Baseline NB | 33 | 116 | 8177 | 1444 |
| Baseline SVM | 1 | 5 | 8288 | 1476 |
| MLC1 | 457 | 1546 | 6747 | 1020 |
| MLC2 | 464 | 1574 | 6719 | 1013 |

**Table 2: Confusion matrix results of the 4 classifiers.**

Recall that the goal of our approach was to better detect patients that will be readmitted within 30 days to the hospital. As we can see from Table 2, we do succeed in this. While the baseline methods NB and SVM only detect respectively 33 and 1 out of 1480 positive patients, our new classifier detects about one third of the patients that will be readmitted within 30 days. Of course, this comes with a higher false positive rate, but this is less problematic than not recognizing patients that will be readmitted within 30 days. If a patient is falsely classified as a patient that will be readmitted within 30 days, this means that the hospital possibly undertakes unnecessary measures for this patient to prevent readmission. These measures will cause additional costs, but they are probably less weighty than costs associated with hospital readmission.

A remarkable conclusion that we can draw from this chart is that the SVM clearly performs worse as baseline method. Although SVM is generally an accurate classifier, it is not able to handle this imbalanced problem well. NB can deal with the imbalanced problem slightly better, but it is only able to detect 2 percent of the patients that will be readmitted within 30 days.

The performances of the two multi-layer classifiers that we proposed do not differ much, probably because the classification in the two first layer are determining for the further final classification. MLC2 is slightly better at detecting patients that will be readmitted within 30 days, but this result is not significant.

**Related Work**

An increasing body of literature attempts to develop and validate the predictive models for risk of hospital readmission. The studies cover readmission due to various diseases (heart failure, pneumonia[10], asthma[11]) and many of them report the outcome for 30 days, though there do exist few models built on different time intervals (60 days[12], 90 days[13], and even 1 year[14]). Each of the developed models exploit different predictor variables and can be classified as using real time data or retrospective data based on the time at which these variables were assessed during an index hospitalization.

One of the significant efforts developed a hierarchical regression model to calculate hospital-specific, risk-standardized, 30-day all-cause readmission rates for Medicare patients hospitalized with heart failure[15]. The model used administrative claims data and focused on primarily cardiovascular and comorbidity variables. The patients used in modeling were limited to the ones more than 65 years old.

In another related work, a real time predictive model is built on the socio-demographic factors of hospitalized heart failure patients to predict the risk of readmission within 30-day time window[16]. Although the model demonstrated good discrimination for 30-day readmission, the dataset size used was much smaller (1372 patients).

In another study, a regression model is developed using Medicare claims along with clinical data of patients discharged between 2004 and 2006[17]. This work focused on patients older than 65 years old and included 24,163 patients from 307 hospitals in their analysis. Our dataset consists of fewer patients but includes more type of data sources.

Another interesting approach develops predictive models for hospital readmission within 30 days that incorporate semantically meaningful derived data elements representing phenotypes[19]. Using this approach, the number of features is reduced drastically, and the data contains less noise. Moreover, clinical knowledge can be introduced into the model and the underlying data representation is abstracted. This preprocessing facilitates the application of data mining algorithms.

To the best of our knowledge, there is only one publication that studies a multi-layer classifier similar to our approach. In this study[18], the authors divide the problem of power transformer fault diagnosis into several sub-problems. The difference with our work is that the authors use the same model for each layer, whereas we propose to use different features and classifiers in each layer.

**Conclusion and Future Work**

In this paper, we introduced a multi-layer classifier to predict if patients discharged from CHS will be readmitted to the hospital within 30 days. Instead of considering this classification as a single task, we subdivide the problem in different subtasks. The advantages of this approach are that we can use different models, feature subsets and training data for each classification subtask, and that the subtasks are more balanced than the original task. An experimental evaluation on a real-world dataset shows that our approach is better at detecting the patients that will be readmitted to the hospital within 30 days than baseline approaches.

In the future we would like to elaborate more on the different models that are used for the subtasks. Currently, we use the same feature selection method in each layer, and we only use two different classifiers over all layers. We want to exploit the fact that the subtasks are smaller classification problems and that we can run more complicated and time-consuming algorithms on them. Moreover, we want study the balance between detecting the patients that will truly be readmitted within 30 days and the cost that is related to the patients that were falsely classified as being readmitted within 30 days.

**Acknowledgements**

The authors would like to thank Multicare Health System for providing anonymized patient data for the analysis.